\documentclass[cameraready]{Interspeech}

\title{Adapting Self-Supervised Speech Representations for\\ Cross-lingual Dysarthria Detection in Parkinson's Disease}

\author[affiliation={1}, correspondingauthor]{Abner}{Hernandez}
\author[affiliation={2}]{Eunjung}{Yeo}
\author[affiliation={2}]{Kwanghee}{Choi}
\author[affiliation={3}]{Chin-Jou}{Li}
\author[affiliation={7}]{Zhengjun}{Yue}
\author[affiliation={8}]{Rohan Kumar}{Das}
\author[affiliation={4}]{Jan}{Rusz}
\author[affiliation={5}]{Mathew}{Magimai Doss}
\author[affiliation={1, 6}]{Juan Rafael}{Orozco-Arroyave}
\author[affiliation={1, 6}]{Tomás}{Arias-Vergara}
\author[affiliation={1}]{Andreas}{Maier}
\author[affiliation={1}]{Elmar}{Nöth}
\author[affiliation={3}]{David R.}{Mortensen}
\author[affiliation={2}]{David}{Harwath}
\author[affiliation={1, 6}]{Paula Andrea}{Perez-Toro}

\address{
    $^1$FAU Erlangen-Nürnberg, Germany, 
    $^2$UT Austin, USA, 
    $^3$CMU, USA, 
    $^4$Czech Technical University in Prague, Czech Republic,
    $^5$Idiap Research Institute, Switzerland,
    $^6$Universidad de Antioquia, Colombia,
    $^7$Shenzhen Loop Area Institute, China,
    $^8$Fortemedia, Singapore
}

\email{abner.hernandez@fau.de}


\usepackage{comment}
\usepackage{hyperref}
\usepackage{cleveref}
\usepackage{multirow}
\usepackage{cite}
\usepackage{subfigure}
\usepackage{tabularray}
\usepackage{booktabs}
\usepackage[HTML]{xcolor}
\usepackage{float}
\usepackage{balance}
\usepackage{bm}

\begin{document}

\maketitle

\begin{abstract}
The limited availability of dysarthric speech data makes cross-lingual detection an important but challenging problem. A key difficulty is that speech representations often encode language-dependent structure that can confound dysarthria detection. We propose a representation-level language shift (LS) that aligns source-language self-supervised speech representations with the target-language distribution using centroid-based vector adaptation estimated from healthy control speech. We evaluate the approach on oral diadochokinesis recordings from Parkinson's disease speech datasets in Czech, German, and Spanish under both cross-lingual and multilingual settings. LS substantially improves sensitivity and F1 in cross-lingual settings, while yielding smaller but consistent gains in multilingual settings. Representation analysis further shows that LS reduces language identity in the embedding space, supporting the interpretation that LS removes language-dependent structure.
\end{abstract}

\section{Introduction}\label{sec:introduction}
Parkinson's disease (PD) is a progressive neurodegenerative disorder that frequently affects motor control, including speech production. A common speech impairment associated with PD is dysarthria~\cite{atalar2023hypokinetic}, a motor speech disorder characterized by abnormalities in respiration, phonation, articulation, resonance, and prosody~\cite{darley1969differential, duffy2019motor}. These speech changes motivate automatic speech analysis methods for PD, yet most prior dysarthria detection studies have been conducted in monolingual settings. Given the global burden of neurological disorders associated with dysarthria~\cite{WHO2024}, developing models that generalize across languages remains an important challenge, motivating recent cross-lingual and multilingual approaches~\cite{orozco2016automatic,yeo2022cross,favaro2023multilingual,kovac2021multilingual}.

However, speech impairments may not manifest identically across languages. A speaker's native language influences phonology, syllable structure, and speech rhythm, which can shape how motor impairments appear in acoustic speech patterns~\cite{liss2013crosslinguistic, kim2017cross, kim2024introduction, yeo2025applications}. Even controlled motor speech tasks may exhibit language-dependent differences. For example, cross-language studies of oral diadochokinesis (DDK) report systematic differences in repetition rate and rhythmic patterns across languages~\cite{kim2025cross}. These findings suggest that language-specific structure may remain present in speech representations and can confound cross-lingual dysarthria detection.


\begin{figure}[t]
    \centering
    \includegraphics[width=\columnwidth]{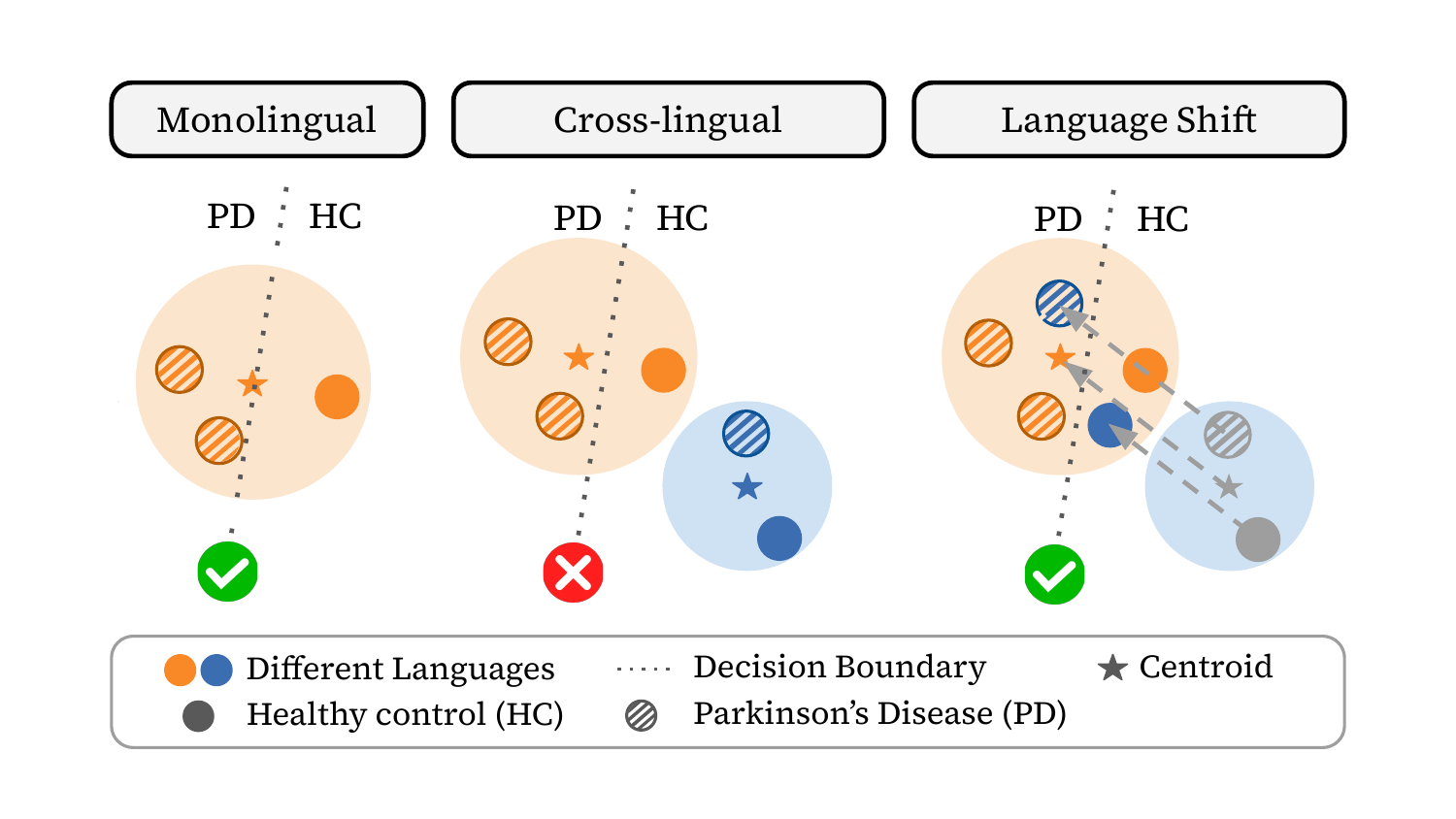}
    \caption{Illustration of the proposed language shift (LS). Source-language HC and PD representations are adapted toward the target-language distribution to reduce language-dependent representation mismatch.
}
    \label{fig:summary}
    \vspace{-3mm}
\end{figure}


These language-dependent differences pose challenges for cross-lingual modeling. Although cross-lingual transfer has shown promise in pathological speech processing~\cite{vasquez2019convolutional, pereztoro22_interspeech, pereztoro23_interspeech}, prior studies also indicate that language-specific representation differences can limit cross-lingual generalization. This highlights the need for methods that mitigate language-dependent variation in speech representations.

From a machine learning perspective, cross-lingual mismatch can be viewed as a form of domain shift. Existing domain adaptation methods, such as correlation alignment (CORAL)~\cite{sun2017correlation}, MMD-based alignment~\cite{cheng2021neural}, and adversarial learning~\cite{Li2018DeepDGA,Tzeng2017AdversarialDDA}, address such shifts. However, many of these approaches require model retraining, additional optimization objectives, or target-domain data, in some cases with labels, which may be limited in clinical speech applications, particularly for dysarthric speech in the target language. Instead of introducing a new trainable domain-adaptation architecture, we examine whether cross-language differences in frozen Self-Supervise Speech (S3M) representations can be approximated by a simple centroid shift estimated from healthy control speech.

This hypothesis is motivated by recent analyses showing that S3M representations encode rich phonetic information \cite{choi2025leveraging} and exhibit structured geometric properties, including approximately linear directions corresponding to phonological features~\cite{choi2026selfb, choi2026self}. If language-specific realization patterns induce systematic shifts in this shared representation space, then aligning source-language embeddings toward the target-language distribution may reduce classifier bias in cross-lingual dysarthria detection. We therefore propose a representation-level language shift (LS)\footnote{Code and experiment scripts are available at: \url{https://github.com/abnerLing/language-shift-dysarthria}.} that estimates language-specific centroids from healthy speech and shifts source-language representations toward the target-language space prior to classification (\Cref{fig:summary}). The method operates directly in representation space and does not require retraining the underlying speech model.

The contributions of this work are threefold:
\begin{itemize}
\item We show that language-dependent structure in DDK representations can confound cross-lingual dysarthria detection.
\item We introduce a simple centroid-based LS that performs adaptation directly in representation space without retraining the speech model.
\item We demonstrate that this representation-level adaptation improves cross-lingual dysarthria detection across Czech, German, and Spanish PD speech datasets.
\end{itemize}


\section{Datasets}\label{sec:datasets}
\begin{table}[t]
\centering
\caption{Demographic and clinical characteristics of the PD datasets in Czech (CZ), German (DE), and Spanish (ES). Values are reported as mean (std. dev.). HC denotes healthy controls.}
\label{tab:dataset}
\footnotesize
\setlength{\tabcolsep}{4pt}
\begin{tabular}{llll}
\toprule
{Lang.} & {Clinical Characteristics} & {PD} & {HC}  \\
\midrule

\multirow{3}{*}{CZ} 
& Gender (F/M) & 20F / 30M & 20F / 30M   \\
& Age (yrs.) & 63.4 (9.5) & 61.6 (11.2)   \\
& Time since diagnosis (yrs.) & 6.7 (4.7) & n/a  \\
& UPDRS-III & 20.1 (10.9) & n/a   \\
& UPDRS-III speech & 0.8 (0.6) & 0.1 (0.3) \\
\midrule

\multirow{3}{*}{DE} 
& Gender (F/M) & 41F / 47M & 44F / 44M  \\
& Age (yrs.) & 66.5 (8.9) & 63.2 (13.9) \\
& Time since diagnosis (yrs.) & 7.1 (5.9) & n/a   \\
& UPDRS-III & 22.7 (10.9) & n/a  \\
& UPDRS-III speech & 1.3 (0.6) & 0.1 (0.3) \\
\midrule

\multirow{3}{*}{ES} 
& Gender (F/M) & 25F / 25M & 25F / 25M    \\
& Age (yrs.) & 61.0 (9.4) & 60.9 (9.4) \\
& Time since diagnosis (yrs.) & 10.6 (9.2) & n/a  \\
& MDS-UPDRS-III & 36.5 (16.5) & n/a   \\
& MDS-UPDRS-III speech & 1.3 (0.8) & 0.2 (0.4) \\

\bottomrule
\end{tabular}
\end{table}
We evaluate our proposed approach on three PD speech datasets in Czech, German (from Germany), and Spanish (the PC-GITA corpus from Colombia~\cite{orozco2014new}). Patients were clinically evaluated using the Unified Parkinson's Disease Rating Scale (UPDRS-III)~\cite{movement2003unified} in the Czech~\cite{rios2024automatic} and German datasets~\cite{bocklet2011detection}, and the Movement Disorder Society revision (MDS-UPDRS-III)~\cite{goetz2008mds} in the Spanish dataset~\cite{perez2021emotional}. A summary of the demographic and clinical characteristics of the datasets is presented in~\Cref{tab:dataset}. 

While the datasets include multiple speech tasks, we constrain our analysis to oral DDK. In this task, speakers repeatedly produce the syllable sequence /pa-ta-ka/, which probes articulatory speed, coordination, and rhythmic control and is widely used in clinical evaluation of dysarthria. We focus on this task because its phonemic content is fixed across recordings, which reduces linguistic variation across languages. Each speaker produced at least one recording of rapid /pa-ta-ka/ repetitions, with some speakers providing two or three recordings.


\section{Method}\label{sec:method}
We investigate cross-language dysarthria detection using self-supervised representations extracted from DDK recordings. Let the \textit{target language} (TGT) denote the language on which the model is evaluated, and \textit{source languages} (SRC) denote the remaining languages that may provide additional training data. Speech embeddings are extracted from pretrained S3Ms and aggregated to obtain speaker-level representations. A logistic regression classifier is trained to distinguish PD from healthy control (HC) speakers. To mitigate language-related variation in the representation space, we introduce a centroid-based language shift that aligns SRC embeddings with the distribution of the TGT prior to classification (\Cref{fig:summary}).


\subsection{Self-supervised Representations}\label{ssec:s3m}
We extract representations from HuBERT-Large~\cite{hsu2021hubert}, WavLM-Large~\cite{chen2022wavlm}, and XLS-R-300M~\cite{babu22_interspeech}. We use the final layer for HuBERT and WavLM and the $12^{\text{th}}$ layer for XLS-R, motivated by prior layer-wise analyses showing that phonological structure varies across S3M architectures and layers~\cite{choi2026self}. The layer choice was also validated separately for each model in our setup. Recordings are processed in non-overlapping chunks of up to 20s, frame-level outputs are mean-pooled within each chunk, chunk-level embeddings are averaged into utterance representations, and utterance representations are averaged into one speaker-level vector.

\subsection{Centroid-Based Language Shift}\label{ssec:centroid}

Recent work has shown that S3M representations encode phonological feature vectors that correspond to approximately linear directions in the embedding space~\cite{choi2026self}. Motivated by this finding, we model cross-language differences as systematic shifts in the representation space and compensate for them using centroid-based vector arithmetic.

Let $\mathbf{x}_{\mathrm{src}} \in \mathbb{R}^d$ denote a representation extracted from a source-language recording. We shift this representation toward the target-language space as:

\begin{equation}
\tilde{\mathbf{x}}_{\mathrm{tgt}}
=
\mathbf{x}_{\mathrm{src}} - \bm{\mu}_{\mathrm{src}} + \bm{\mu}_{\mathrm{tgt}},
\label{eq:shift}
\end{equation}

where $\bm{\mu}_{\mathrm{src}}$ and $\bm{\mu}_{\mathrm{tgt}}$ are the centroids of the source and target languages, respectively. Intuitively, this operation subtracts the average representation of the source language and replaces it with that of the target language (illustrated by the dotted arrow in \Cref{fig:summary}).

We compute the centroid using only HC speakers:
\begin{equation}
\bm\mu_{\ell}
=
\frac{1}{N_{\ell,\mathrm{HC}}}
\sum_{i \in \mathrm{HC}_{\ell}^{\mathrm{train}}}
\mathbf{x}_i,
\label{eq:hc}
\end{equation}
where $N_{\ell,\mathrm{HC}}$ is the number of training HC speakers in language $\ell \in \{\mathrm{src}, \mathrm{tgt}\}$.
Language centroids are estimated within each cross-validation fold to avoid test-set leakage.

Using HC speakers isolates language-specific variation from pathology-related deviations, allowing the transformation to act as a language normalization step prior to dysarthria classification.
Further, as our analysis focuses on DDK recordings, where the phonemic content (/pa-ta-ka/) is fixed across languages, the centroid shift in \cref{eq:shift} can be interpreted as regularizing these language-dependent representation shifts.


\section{Experiments}\label{sec:experiments}
\begin{table*}[t]
\centering
\caption{Cross-lingual dysarthria detection results before and after applying the proposed language shift (LS). Each metric is reported as \textit{Cross. / Ours}, where Cross. denotes the baseline without LS and Ours denotes results after applying LS. \textbf{Bold} indicates the better value between the two settings. \textbf{\textcolor{blue}{Blue}} indicates improvements with non-overlapping confidence intervals.}
\label{tab:main-results-crosslingual}
\small

\begin{tabular}{@{}llccc@{}}
\toprule
Lang. & S3M & Spec. (Cross.\ / Ours) & Sens. (Cross.\ / Ours) & F1 (Cross.\ / Ours) \\
\midrule
\multirow{3}{*}{CZ}
 & HuBERT & \textbf{\textcolor{blue}{0.98}}{\scriptsize$\pm$0.02} / 0.43{\scriptsize$\pm$0.10} & 0.35{\scriptsize$\pm$0.11} / \textbf{\textcolor{blue}{0.93}}{\scriptsize$\pm$0.05} & 0.48{\scriptsize$\pm$0.10} / \textbf{\textcolor{blue}{0.74}}{\scriptsize$\pm$0.04} \\
 & WavLM & \textbf{\textcolor{blue}{0.97}}{\scriptsize$\pm$0.03} / 0.32{\scriptsize$\pm$0.12} & 0.26{\scriptsize$\pm$0.08} / \textbf{\textcolor{blue}{0.90}}{\scriptsize$\pm$0.05} & 0.38{\scriptsize$\pm$0.10} / \textbf{\textcolor{blue}{0.70}}{\scriptsize$\pm$0.03} \\
 & XLS-R & \textbf{\textcolor{blue}{0.80}}{\scriptsize$\pm$0.10} / 0.31{\scriptsize$\pm$0.10} & 0.59{\scriptsize$\pm$0.12} / \textbf{\textcolor{blue}{0.81}}{\scriptsize$\pm$0.09} & 0.64{\scriptsize$\pm$0.08} / \textbf{0.65}{\scriptsize$\pm$0.04} \\
\midrule
\multirow{3}{*}{DE}
 & HuBERT & \textbf{\textcolor{blue}{0.90}}{\scriptsize$\pm$0.04} / 0.54{\scriptsize$\pm$0.07} & 0.28{\scriptsize$\pm$0.06} / \textbf{\textcolor{blue}{0.65}}{\scriptsize$\pm$0.09} & 0.39{\scriptsize$\pm$0.07} / \textbf{\textcolor{blue}{0.61}}{\scriptsize$\pm$0.06} \\
 & WavLM & \textbf{\textcolor{blue}{0.87}}{\scriptsize$\pm$0.03} / 0.48{\scriptsize$\pm$0.06} & 0.28{\scriptsize$\pm$0.07} / \textbf{\textcolor{blue}{0.67}}{\scriptsize$\pm$0.06} & 0.39{\scriptsize$\pm$0.08} / \textbf{\textcolor{blue}{0.61}}{\scriptsize$\pm$0.04} \\
 & XLS-R & \textbf{\textcolor{blue}{0.79}}{\scriptsize$\pm$0.06} / 0.40{\scriptsize$\pm$0.07} & 0.60{\scriptsize$\pm$0.08} / \textbf{\textcolor{blue}{0.75}}{\scriptsize$\pm$0.07} & \textbf{0.65}{\scriptsize$\pm$0.05} / 0.64{\scriptsize$\pm$0.04} \\
\midrule
\multirow{3}{*}{ES}
 & HuBERT & \textbf{\textcolor{blue}{0.99}}{\scriptsize$\pm$0.02} / 0.61{\scriptsize$\pm$0.10} & 0.29{\scriptsize$\pm$0.06} / \textbf{\textcolor{blue}{0.83}}{\scriptsize$\pm$0.09} & 0.44{\scriptsize$\pm$0.07} / \textbf{\textcolor{blue}{0.74}}{\scriptsize$\pm$0.05} \\
 & WavLM & \textbf{\textcolor{blue}{0.96}}{\scriptsize$\pm$0.03} / 0.53{\scriptsize$\pm$0.11} & 0.20{\scriptsize$\pm$0.07} / \textbf{\textcolor{blue}{0.83}}{\scriptsize$\pm$0.09} & 0.30{\scriptsize$\pm$0.10} / \textbf{\textcolor{blue}{0.71}}{\scriptsize$\pm$0.04} \\
 & XLS-R & \textbf{\textcolor{blue}{0.80}}{\scriptsize$\pm$0.08} / 0.46{\scriptsize$\pm$0.10} & 0.55{\scriptsize$\pm$0.09} / \textbf{\textcolor{blue}{0.85}}{\scriptsize$\pm$0.08} & 0.61{\scriptsize$\pm$0.06} / \textbf{0.70}{\scriptsize$\pm$0.04} \\
\bottomrule
\end{tabular}

\end{table*}

\begin{table*}[t]
\centering
\caption{Multilingual dysarthria detection results before and after applying the proposed language shift (LS). Mono. and Multi. denote results without LS, while Ours denotes results after applying LS. \textbf{Bold} indicates the best value among the three settings. \textbf{\textcolor{blue}{Blue}} indicates improvements with non-overlapping confidence intervals.}
\label{tab:main-results-multilingual}
\small
\begin{tabular}{@{}llccc@{}}
\toprule
Lang. & S3M & Specificity (Mono.\ / Multi.\ / Ours) & Sensitivity (Mono.\ / Multi.\ / Ours) & F1 (Mono.\ / Multi.\ / Ours) \\
\midrule
\multirow{3}{*}{CZ}
 & HuBERT & 0.53{\scriptsize$\pm$0.12} / 0.59{\scriptsize$\pm$0.10} / \textbf{0.66}{\scriptsize$\pm$0.10} & \textbf{0.89}{\scriptsize$\pm$0.07} / 0.83{\scriptsize$\pm$0.05} / 0.83{\scriptsize$\pm$0.07} & 0.76{\scriptsize$\pm$0.05} / 0.74{\scriptsize$\pm$0.04} / \textbf{0.76}{\scriptsize$\pm$0.04} \\
 & WavLM & 0.55{\scriptsize$\pm$0.11} / \textbf{0.58}{\scriptsize$\pm$0.11} / \textbf{0.58}{\scriptsize$\pm$0.10} & \textbf{0.91}{\scriptsize$\pm$0.05} / 0.85{\scriptsize$\pm$0.05} / 0.83{\scriptsize$\pm$0.08} & \textbf{0.77}{\scriptsize$\pm$0.03} / 0.75{\scriptsize$\pm$0.03} / 0.74{\scriptsize$\pm$0.05} \\
 & XLS-R & 0.31{\scriptsize$\pm$0.09} / 0.47{\scriptsize$\pm$0.13} / \textbf{0.48}{\scriptsize$\pm$0.07} & \textbf{\textcolor{blue}{0.93}}{\scriptsize$\pm$0.03} / 0.77{\scriptsize$\pm$0.09} / 0.81{\scriptsize$\pm$0.07} & \textbf{0.71}{\scriptsize$\pm$0.03} / 0.67{\scriptsize$\pm$0.06} / 0.69{\scriptsize$\pm$0.04} \\
\midrule
\multirow{3}{*}{DE}
 & HuBERT & 0.47{\scriptsize$\pm$0.06} / 0.48{\scriptsize$\pm$0.06} / \textbf{0.49}{\scriptsize$\pm$0.05} & \textbf{0.91}{\scriptsize$\pm$0.03} / 0.84{\scriptsize$\pm$0.06} / 0.89{\scriptsize$\pm$0.05} & \textbf{0.75}{\scriptsize$\pm$0.02} / 0.71{\scriptsize$\pm$0.04} / 0.74{\scriptsize$\pm$0.02} \\
 & WavLM & 0.39{\scriptsize$\pm$0.05} / \textbf{0.40}{\scriptsize$\pm$0.05} / 0.39{\scriptsize$\pm$0.07} & \textbf{0.90}{\scriptsize$\pm$0.04} / 0.87{\scriptsize$\pm$0.05} / 0.86{\scriptsize$\pm$0.07} & \textbf{0.72}{\scriptsize$\pm$0.02} / 0.70{\scriptsize$\pm$0.04} / 0.69{\scriptsize$\pm$0.03} \\
 & XLS-R & 0.41{\scriptsize$\pm$0.06} / 0.45{\scriptsize$\pm$0.07} / \textbf{0.46}{\scriptsize$\pm$0.06} & \textbf{0.90}{\scriptsize$\pm$0.04} / 0.86{\scriptsize$\pm$0.06} / 0.85{\scriptsize$\pm$0.05} & \textbf{0.72}{\scriptsize$\pm$0.02} / 0.71{\scriptsize$\pm$0.03} / 0.71{\scriptsize$\pm$0.03} \\
\midrule
\multirow{3}{*}{ES}
 & HuBERT & 0.41{\scriptsize$\pm$0.09} / 0.50{\scriptsize$\pm$0.10} / \textbf{0.57}{\scriptsize$\pm$0.09} & \textbf{0.88}{\scriptsize$\pm$0.06} / \textbf{0.88}{\scriptsize$\pm$0.06} / 0.85{\scriptsize$\pm$0.07} & 0.71{\scriptsize$\pm$0.04} / 0.73{\scriptsize$\pm$0.03} / \textbf{0.74}{\scriptsize$\pm$0.04} \\
 & WavLM & 0.31{\scriptsize$\pm$0.10} / 0.52{\scriptsize$\pm$0.08} / \textbf{0.54}{\scriptsize$\pm$0.09} & \textbf{0.88}{\scriptsize$\pm$0.06} / 0.87{\scriptsize$\pm$0.07} / \textbf{0.88}{\scriptsize$\pm$0.06} & 0.68{\scriptsize$\pm$0.02} / 0.73{\scriptsize$\pm$0.03} / \textbf{0.75}{\scriptsize$\pm$0.04} \\
 & XLS-R & 0.35{\scriptsize$\pm$0.09} / 0.61{\scriptsize$\pm$0.08} / \textbf{0.62}{\scriptsize$\pm$0.09} & \textbf{0.91}{\scriptsize$\pm$0.04} / 0.81{\scriptsize$\pm$0.10} / 0.80{\scriptsize$\pm$0.09} & 0.70{\scriptsize$\pm$0.02} / 0.72{\scriptsize$\pm$0.06} / \textbf{0.72}{\scriptsize$\pm$0.04} \\
\bottomrule
\end{tabular}

\end{table*}

\subsection{Experimental settings}\label{ssec:settings}
We evaluate dysarthria detection under two settings depending on whether PD speech from TGT is available during training: (1) without TGT PD (\Cref{sssec:xling}), and (2) with TGT PD (\Cref{sssec:mling}).
In both settings, we assume that HC and PD speech from the SRC1 and SRC2, as well as HC speech from TGT, are available during training.

Speaker-level classification uses logistic regression with 5-fold stratified cross-validation, treating one language as the target and the others as sources. 
Motivated by the use of high-sensitivity operating points in clinical screening, where missed cases are particularly undesirable~\cite{maxim2014screening,smits2010youden}, decision thresholds are selected using nested cross-validation to enforce a minimum sensitivity of 0.9, choosing the highest specificity among valid thresholds. The same threshold-selection rule is applied uniformly to all baselines and LS variants, therefore, performance differences reflect changes in the representations rather than method-specific threshold tuning. Experiments are evaluated using mean F1-score, sensitivity, and specificity across folds and 3 random seeds.

To control class imbalance, the number of speakers per class is capped to the smallest class size across datasets, ensuring balanced HC and PD contributions during training. All experiments use speaker-independent train–test splits.

\begin{figure*}[t!]
\centering
\includegraphics[width=0.95\textwidth]{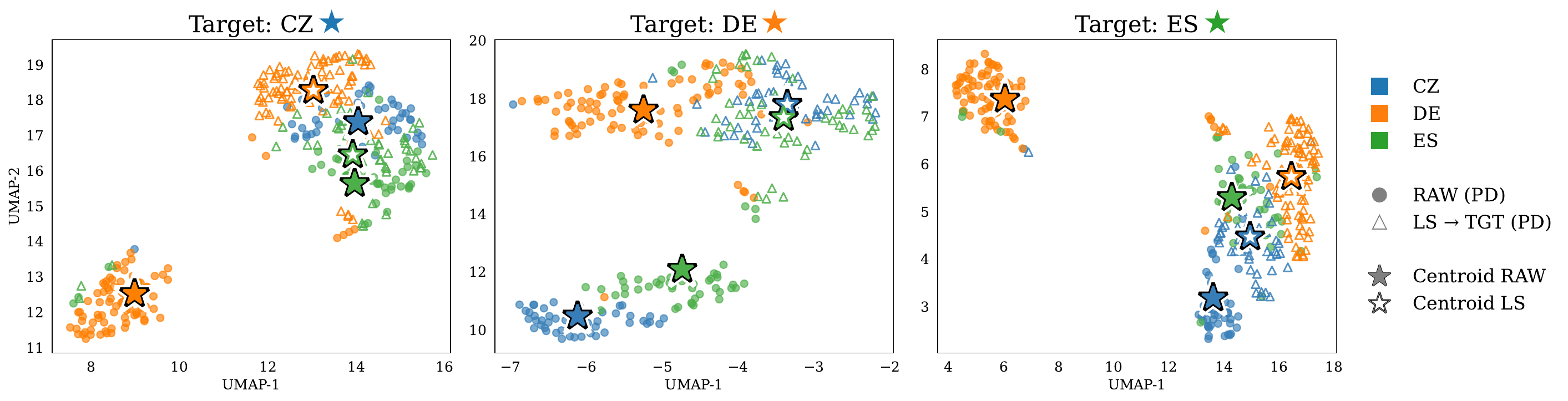}
\caption{UMAP visualization of HuBERT PD speaker representations with and without shifting. Circles denote raw source representations and triangles show representations shifted to the target-language (LS$\rightarrow$TGT).}
\label{fig:umap_es-de-cz}
\vspace{-2mm}
\end{figure*}

\subsection{Cross-Lingual Transfer with Language Shift}\label{sssec:xling}
\textbf{Settings.}
This setup simulates a scenario where PD speech from TGT is unavailable during training. 
Therefore, the model must rely on cross-lingual transfer to learn HC–PD separation.
Specifically, the model is trained using HC and PD speakers from SRC1 and SRC2, and HC speakers from TGT. 
Evaluation is performed on TGT using both HC and PD.

\noindent\textbf{Baseline.}
We compare our method with a \textit{cross-lingual} (Cross.) baseline. 
For both the baseline and ours, the classifier is trained using HC data from SRC1, SRC2, and TGT, and PD data from SRC1 and SRC2.

\noindent\textbf{Results.}
\Cref{tab:main-results-crosslingual} shows that cross-lingual transfer without language shifting produces highly imbalanced predictions. Models tend to achieve very high specificity but low sensitivity, indicating that PD speakers are frequently misclassified as healthy controls. For instance, HuBERT reaches specificity of 0.98, 0.90, and 0.99 for Czech, German, and Spanish, while sensitivity drops to 0.35, 0.28, and 0.29, respectively.

This imbalance suggests that language-related representation differences interfere with pathology classification (\Cref{fig:summary}). Because training data contain PD speakers only from the source languages, the model may implicitly rely on language-dependent patterns rather than pathology cues when evaluated on target-language PD speakers.

Applying the proposed language shift alleviates this issue. After shifting source-language representations toward the target-language centroid, sensitivity increases substantially across all models and languages. For HuBERT, F1 improves to 0.74, 0.61, and 0.74 on Czech, German, and Spanish, respectively, with sensitivity recovering to 0.93, 0.65, and 0.83. Similar improvements are observed for WavLM and XLS-R, although German remains the most challenging target language.

\subsection{Multilingual Transfer with Language Shift}\label{sssec:mling}
\textbf{Settings.}
Compared to \Cref{sssec:xling}, this setup considers a more relaxed scenario where PD speech from TGT is available during training. 
The model is trained using HC and PD from all SRC1, SRC2, and TGT.
Same with \Cref{sssec:xling}, evaluation is performed on TGT using both HC and PD.

\noindent\textbf{Baseline.}
We compare ours with two baselines: \textit{monolingual} (mono.) and \textit{multilingual} (multi.).
In the monolingual setting, the classifier is trained using only HC and PD of TGT. 
In the multilingual setting, the classifier is trained using HC and PD from all SRC1, SRC2, and TGT.

\noindent\textbf{Results.}
\Cref{tab:main-results-multilingual} presents results when PD recordings from the target language are available during training. In this multilingual setting, the severe sensitivity–specificity imbalance observed in the cross-lingual experiments largely disappears, as the classifier can directly learn pathological cues in the target language.

Consequently, the differences between monolingual, multilingual, and proposed method are smaller, with most F1 variations falling within overlapping confidence intervals. Nevertheless, applying LS often improves specificity without substantially degrading sensitivity. For example, with HuBERT, specificity increases from 0.59 to 0.66 for Czech and from 0.50 to 0.57 for Spanish while maintaining similar sensitivity levels.

Overall, these results indicate that the proposed language shift is most beneficial in the cross-lingual setting where target-language PD data are unavailable. When PD data from the target language are included during training, both monolingual and multilingual models already capture much of the language-specific structure, and the additional gains from LS become more modest.

\section{Discussion}\label{sec:analysis}
We analyze the representation space to better understand why LS improves cross-lingual dysarthria detection. Specifically, we examine whether S3M representations encode language-dependent structure even for the controlled DDK task, and how LS alters this structure. A qualitative overview using HuBERT is shown in\Cref{fig:umap_es-de-cz}.

\noindent\textbf{Language distances in representation space.}
Before applying LS, Czech and Spanish are closest in S3M space ($L_2=1.47$), whereas German is more distant from both (CZ: $2.56$, ES: $2.49$). This larger separation may help explain why German shows smaller gains from LS in the cross-lingual experiments: while a centroid shift compensates for average language offsets, larger distributional differences may require more complex transformations. The distances are computed as the Euclidean distance between per-language centroids in the original 1024-dimensional S3M space, \textit{e.g.}, $L_2(\text{CZ}, \text{ES}) = ||\mu_{\text{CZ}} - \mu_{\text{ES}}||$. 

These patterns may partly reflect linguistic differences between the languages~\cite{ramus1999correlates}. For example, German is typically classified as stress-timed, whereas Czech and Spanish are more syllable-timed~\cite{nespor201149}, which may influence temporal patterns in DDK production. Cross-language evidence suggests that speakers of stress-timed and syllable-timed languages can differ in DDK timing variability~\cite{kim2025cross}, potentially contributing to the observed centroid structure.

\noindent\textbf{Language identity in S3M representations.}
To examine how LS affects language-dependent structure in the representations, we train a linear support vector machine (SVM) with default regularization ($C=1$) on HC representations to predict language identity. Input features are z-normalized using statistics from the HC training set, and the classifier is evaluated on PD speakers. Without LS, the probe achieves 96\% accuracy, confirming that language identity is strongly encoded in S3M representations even for dysarthric speech. After applying LS toward any target language, accuracy drops to near chance level (34\%, 33\%, and 29\% for Czech, German, and Spanish, respectively). This suggests that the proposed shift effectively removes language-dependent structure in the representation space, reducing cues that could otherwise confound cross-lingual pathology classification.

\section{Conclusion}\label{sec:conclusion}
This work shows that even controlled motor speech tasks such as DDK contain language-dependent structure in self-supervised speech representations, which can hinder cross-lingual dysarthria detection. To address this, we proposed a simple centroid-based language shift that aligns source-language representations with the target-language distribution directly in representation space.

Experiments on Czech, German, and Spanish PD datasets demonstrate that this shift consistently improves cross-lingual detection, substantially increasing sensitivity and F1 scores across models and target languages. In multilingual settings, the gains are smaller but often improve specificity, suggesting that access to target-language PD data already captures much of the language-specific variation. Overall, these findings highlight that language effects persist and should be considered when designing cross-lingual clinical speech models. 

One limitation of this study is that each language is drawn from a different dataset, making it difficult to disentangle linguistic effects from corpus-specific characteristics. In addition, our analysis focuses on oral DDK recordings, which represent a highly controlled speech task. Future work should therefore extend the analysis to less controlled and more natural speech settings.

\section{Generative AI Use Disclosure}
Generative artificial intelligence tools were used to assist with language editing, clarity of presentation, and code drafting/debugging. All research ideas, methodology, experiments, analyses, and interpretations were conceived, verified, and carried out by the authors, who take full responsibility for the originality, validity, and integrity of the work.

\bibliographystyle{IEEEtran}
\bibliography{mybib}

@article{kim2025cross,
  title={{A Cross-Language Study of Oral Diadochokinesis: Rates and Rhythm}},
  author={Kim, Yunjung and Berry, Jeffrey and Lee, Seung Jin and Lin, Lifeng},
  journal={Folia Phoniatrica et Logopaedica},
  year={2025}
}

@article{darley1969differential,
  title={{Differential diagnostic patterns of dysarthria}},
  author={Darley, F. L. and Aronson, A. E. and Brown, J. R.},
  journal={Journal of speech and hearing research},
  volume={12},
  number={2},
  year={1969},
  publisher={American Speech-Language-Hearing Association}
}

@article{kim2024introduction,
  title={{Introduction to the Forum: Native Language, Dialect, and Foreign Accent in Dysarthria}},
  author={Kim, Yunjung},
  journal={Journal of Speech, Language, and Hearing Research},
  volume={67},
  number={9},
  year={2024},
  publisher={American Speech-Language-Hearing Association}
}

@inproceedings{orozco2014new,
  title={New {Spanish} speech corpus database for the analysis of people suffering from Parkinson's disease.},
  author={Orozco-Arroyave, Juan Rafael and Arias-Londo{\~n}o, Juli{\'a}n David and Vargas-Bonilla, Jes{\'u}s Francisco and others},
  booktitle={Proc. of LREC},
  pages={342--347},
  year={2014}
}

@inproceedings{perez2021emotional,
  title={{Emotional state modeling for the assessment of depression in Parkinson’s disease}},
  author={P{\'e}rez-Toro, Paula Andrea and V{\'a}squez-Correa, Juan Camilo and Arias-Vergara, Tom{\'a}s and Klumpp, Philipp and Schuster, Maria and N{\"o}th, Elmar and Orozco-Arroyave, Juan Rafael},
  booktitle={Proc. of International Conference on Text, Speech, and Dialogue},
  pages={457--468},
  year={2021},
  organization={Springer}
}

@article{rios2024automatic,
  title={{Automatic speech-based assessment to discriminate Parkinson’s disease from essential tremor with a cross-language approach}},
  author={Rios-Urrego, Cristian David and Rusz, Jan and Orozco-Arroyave, Juan Rafael},
  journal={npj Digital Medicine},
  volume={7},
  number={1},
  pages={37},
  year={2024},
  publisher={Nature Publishing Group UK London}
}

@inproceedings{bocklet2011detection,
  title={{Detection of persons with Parkinson's disease by acoustic, vocal, and prosodic analysis}},
  author={Bocklet, Tobias and N{\"o}th, Elmar and Stemmer, Georg and Ruzickova, Hana and Rusz, Jan},
  booktitle={Proc. of ASRU},
  pages={478--483},
  year={2011},
}

@article{yeo2025applications,
  title={{Applications of artificial intelligence for cross-language intelligibility assessment of dysarthric speech}},
  author={Yeo, Eunjung and Liss, Julie M and Berisha, Visar and Mortensen, David R},
  journal={Perspectives of the ASHA Special Interest Groups},
  volume={10},
  number={6},
  pages={2298--2308},
  year={2025},
  publisher={American Speech-Language-Hearing Association}
}

@inproceedings{vasquez2019convolutional,
  title={{Convolutional neural networks and a transfer learning strategy to classify Parkinson’s disease from speech in three different languages}},
  author={V{\'a}squez-Correa, Juan Camilo and Arias-Vergara, Tomas and Rios-Urrego, Cristian D and Schuster, Maria and Rusz, Jan and Orozco-Arroyave, Juan Rafael and N{\"o}th, Elmar},
  booktitle={Iberoamerican Congress on Pattern Recognition},
  pages={697--706},
  year={2019},
  organization={Springer}
}

@inproceedings{pereztoro22_interspeech,
  title     = {{Alzheimer's Detection from English to Spanish Using Acoustic and Linguistic Embeddings}},
  author    = {Paula Andrea Pérez-Toro and Philipp Klumpp and Abner Hernandez and Tomas Arias and Patricia Lillo and Andrea Slachevsky and Adolfo Martín García and Maria Schuster and Andreas K. Maier and Elmar Noeth and Juan Rafael Orozco-Arroyave},
  year      = {2022},
  booktitle = {{Proc. of Interspeech}},
  pages     = {2483--2487},
  doi       = {10.21437/Interspeech.2022-10883},
  issn      = {2958-1796},
}

@inproceedings{pereztoro23_interspeech,
  title     = {{Automatic Assessment of Alzheimer's across Three Languages Using Speech and Language Features}},
  author    = {Paula A. Pérez-Toro and Tomás Arias-Vergara and Franziska Braun and Florian Hönig and Carlos A. Tobón-Quintero and David Aguillón and Francisco Lopera and Liliana Hincapié-Henao and Maria Schuster and Korbinian Riedhammer and Andreas Maier and Elmar Nöth and Juan Rafael Orozco-Arroyave},
  year      = {2023},
  booktitle = {{Proc. of Interspeech}},
  pages     = {1748--1752},
  doi       = {10.21437/Interspeech.2023-2079},
  issn      = {2958-1796},
}

@article{movement2003unified,
  title={{The unified Parkinson's disease rating scale (UPDRS): status and recommendations}},
  author={Movement Disorder Society Task Force on Rating Scales for Parkinson's Disease},
  journal={Movement Disorders},
  volume={18},
  number={7},
  pages={738--750},
  year={2003},
  publisher={Wiley Online Library}
}

@article{goetz2008mds,
  title={{MDS-Unified Parkinson’s Disease Rating Scale (MDS-UPDRS)}},
  author={Goetz, Christopher G and Tilley, Barbara and Shaftman, Stephanie R and Stebbins, Glenn T and Fahn, Stanley and Martinez-Martin, Pablo and Poewe, Werner and Sampaio, Cristina and Stern, Matthew B and Dodel, Richard and others},
  journal={International Parkinson and Movement Disorder Society},
  year={2008},
  url={https://www.movementdisorders.org/MDS-Files1/PDFs/Rating-Scales/MDS-UPDRS_English_FINAL.pdf}
}

@article{hsu2021hubert,
  title={{Hubert: Self-supervised speech representation learning by masked prediction of hidden units}},
  author={Hsu, Wei-Ning and Bolte, Benjamin and Tsai, Yao-Hung Hubert and Lakhotia, Kushal and Salakhutdinov, Ruslan and Mohamed, Abdelrahman},
  journal={IEEE/ACM transactions on audio, speech, and language processing},
  volume={29},
  pages={3451--3460},
  year={2021},
}

@article{chen2022wavlm,
  title={{WavLM: Large-Scale Self-Supervised Pre-Training for Full Stack Speech Processing}},
  author={Chen, Sanyuan and Wang, Chengyi and Chen, Zhengyang and Wu, Yu and Liu, Shujie and Chen, Zhuo and Li, Jinyu and Kanda, Naoyuki and Yoshioka, Takuya and Xiao, Xiong and others},
  journal={Journal of Selected Topics in Signal Processing},
  volume={16},
  number={6},
  pages={1505--1518},
  year={2022},
}

@inproceedings{babu22_interspeech,
  title     = {{XLS-R: Self-supervised Cross-lingual Speech Representation Learning at Scale}},
  author    = {Arun Babu and Changhan Wang and Andros Tjandra and Kushal Lakhotia and Qiantong Xu and Naman Goyal and Kritika Singh and Patrick {von Platen} and Yatharth Saraf and Juan Pino and Alexei Baevski and Alexis Conneau and Michael Auli},
  year      = {2022},
  booktitle = {{Proc. of Interspeech}},
  pages     = {2278--2282},
  doi       = {10.21437/Interspeech.2022-143},
  issn      = {2958-1796},
}

@article{choi2026selfb,
  title= {{[b]=[d]-[t]+[p]: Self-supervised Speech Models Discover Phonological Vector Arithmetic}},
  author={Choi, Kwanghee and Yeo, Eunjung and Cho, Cheol Jun and others},
  journal={Proc. of ACL Findings},
  year={2026}
}

@article{choi2026self,
  title={Self-supervised speech models encode phonetic context via position-dependent orthogonal subspaces},
  author={Choi, Kwanghee and Yeo, Eunjung and Cho, Cheol Jun and Mortensen, David R and Harwath, David},
  journal={arXiv preprint arXiv:2603.12642},
  year={2026}
}

@article{liss2013crosslinguistic,
  title={{Crosslinguistic application of English-centric rhythm descriptors in motor speech disorders}},
  author={Liss, Julie M and Utianski, Rene and Lansford, Kaitlin},
  journal={Folia Phoniatrica et Logopaedica},
  volume={65},
  number={1},
  year={2013},
  publisher={S. Karger AG Basel, Switzerland}
}

@article{kim2017cross,
  title={{A cross-language study of acoustic predictors of speech intelligibility in individuals with Parkinson's disease}},
  author={Kim, Yunjung and Choi, Yaelin},
  journal={Journal of speech and hearing research},
  volume={60},
  number={9},
  year={2017},
  publisher={American Speech-Language-Hearing Association}
}

@inproceedings{choi2025leveraging,
  title={{Leveraging allophony in self-supervised speech models for atypical pronunciation assessment}},
  author={Choi, Kwanghee and Yeo, Eunjung and Chang, Kalvin and Watanabe, Shinji and Mortensen, David R},
  booktitle={Proc. of NAACL},
  pages={2613--2628},
  year={2025}
}

@article{ramus1999correlates,
  title={{Correlates of linguistic rhythm in the speech signal}},
  author={Ramus, Franck and Nespor, Marina and Mehler, Jacques},
  journal={Cognition},
  volume={73},
  number={3},
  pages={265--292},
  year={1999},
  publisher={Elsevier}
}

@misc{WHO2024,
  author = {{World Health Organization}},
  title = {Over 1 in 3 people affected by neurological conditions, the leading cause of illness and disability worldwide},
  year = {2024},
}

@article{favaro2023multilingual,
  title={{Multilingual evaluation of interpretable biomarkers to represent language and speech patterns in Parkinson's disease}},
  author={Favaro, Anna and Moro-Vel{\'a}zquez, Laureano and Butala, Ankur and Motley, Chelsie and Cao, Tianyu and Stevens, Robert David and Villalba, Jes{\'u}s and Dehak, Najim},
  journal={Frontiers in Neurology},
  volume={14},
  year={2023},
}

@article{nespor201149,
  title={{49 Stress-timed vs. Syllable-timed Languages}},
  author={Nespor, Marina and Shukla, Mohinish and Mehler, Jacques},
  year={2011}
}

@article{atalar2023hypokinetic,
  title={{Hypokinetic Dysarthria in Parkinson's Disease: A Narrative Review}},
  author={Atalar, Merve Sapmaz and Oguz, Ozlem and Genc, Gencer},
  journal={Medical Bulletin of Sisli Etfal Hospital},
  volume={57},
  number={2},
  year={2023}
}

@book{duffy2019motor,
  author    = {Duffy, Joseph R.},
  title     = {Motor Speech Disorders: Substrates, Differential Diagnosis, and Management},
  edition   = {4},
  publisher = {Elsevier},
  year      = {2020}
}

@inproceedings{yeo2022cross,
  title={{Cross-lingual dysarthria severity classification for English, Korean, and Tamil}},
  author={Yeo, Eun Jung and Choi, Kwanghee and Kim, Sunhee and Chung, Minhwa},
  booktitle={Proc. of APSIPA},
  pages={566--574},
  year={2022},
}

@article{orozco2016automatic,
  title={{Automatic detection of Parkinson's disease in running speech spoken in three different languages}},
  author={Orozco-Arroyave, Juan Rafael and H{\"o}nig, F and Arias-Londo{\~n}o, JD and Vargas-Bonilla, JF and Daqrouq, K and Skodda, S and Rusz, J and N{\"o}th, E},
  journal={The Journal of the Acoustical Society of America},
  volume={139},
  number={1},
  pages={481--500},
  year={2016},
  publisher={AIP Publishing}
}

@article{cheng2021neural,
  title={Neural tangent kernel maximum mean discrepancy},
  author={Cheng, Xiuyuan and Xie, Yao},
  journal={Advances in Neural Information Processing Systems},
  volume={34},
  pages={6658--6670},
  year={2021}
}

@incollection{sun2017correlation,
  title={Correlation alignment for unsupervised domain adaptation},
  author={Sun, Baochen and Feng, Jiashi and Saenko, Kate},
  booktitle={Domain adaptation in computer vision applications},
  pages={153--171},
  year={2017},
  publisher={Springer}
}

@inproceedings{kovac2021multilingual,
  title={Multilingual analysis of speech and voice disorders in patients with parkinson’s disease},
  author={Kovac, Daniel and Mekyska, Jiri and Galaz, Zoltan and Brabenec, Lubos and Kostalova, Milena and Rapcsak, Steven Z and Rektorova, Irena},
  booktitle={Proc. of International conference on Telecommunications and Signal Processing (TSP)},
  pages={273--277},
  year={2021},
}

@inproceedings{Li2018DeepDGA,
  title={Deep Domain Generalization via Conditional Invariant Adversarial Networks},
  author={Ya Li and Xinmei Tian and Mingming Gong and Yajing Liu and Tongliang Liu and Kun Zhang and D. Tao},
  booktitle={European Conference on Computer Vision},
  year={2018},
}

@article{Tzeng2017AdversarialDDA,
  title={Adversarial Discriminative Domain Adaptation},
  author={Eric Tzeng and Judy Hoffman and Kate Saenko and Trevor Darrell},
  journal={Proc. of CVPR},
  year={2017},
  pages={2962-2971},
}

@article{maxim2014screening,
  title={Screening tests: a review with examples},
  author={Maxim, L Daniel and Niebo, Ron and Utell, Mark J},
  journal={Inhalation toxicology},
  volume={26},
  number={13},
  pages={811--828},
  year={2014},
  publisher={Taylor \& Francis}
}

@article{smits2010youden,
  title={A note on Youden's J and its cost ratio},
  author={Smits, Niels},
  journal={BMC medical research methodology},
  volume={10},
  number={1},
  pages={89},
  year={2010},
  publisher={Springer}
}

\end{document}